# AVOIDER ROBOT DESIGN TO DIM THE FIRE WITH DT-BASIC MINI SYSTEM

Eri Prasetyo, Wahyu K.R., Bumi Prabu Prabowo
Electrical Engineering, Gunadarma University
Jakarta, Indonesia
E-mail : eri@staff.gunadarma.ac.id, wahyukr@staff.gunadarma.ac.id , Kbumy_fr@plasa.com

*Abstract –* **Avoider robot is mean robot who is designed to avoid the block in around. Except that, this robot is also added by an addition aplication to dim the fire. This robot is made with ultrasonic sensor PING. This sensor is set on the front, right and left from robot. This sensor is used robot to look for the right street, so that robot can walk on. After the robot can look for the right street, next accomplished the robot is looking for the fire in around. And the next, dim the fire with fan. This robot is made with basic stamp 2 microcontroller. And that microcontroller can be found in dt-basic mini system module. This robot is made with servo motor on the right and left side, which is used to movement.**

# AVOIDER ROBOT DESIGN TO DIM THE FIRE WITH DT-BASIC MINI SYSTEM

*Abstract* – Avoider robot is mean robot who is designed to avoid the block in around. Except that, this robot is also added by an addition aplication to dim the fire. This robot is made with ultrasonic sensor PING. This sensor is set on the front, right and left from robot. This sensor is used robot to look for the right street, so that robot can walk on. After the robot can look for the right street, next accomplished the robot is looking for the fire in around. And the next, dim the fire with fan. This robot is made with basic stamp 2 microcontroller. And that microcontroller can be found in dt-basic mini system module. This robot is made with servo motor on the right and left side, which is used to movement.

## 1. INTRODUCTION

Currently, electronics technology already grow up. This situation make us easy to get microcontroller kit with cheap price. And now, to make possible us make a robot. And especially, now every year is made a competition robot. That reason, whom make us to thing to make avoider robot.

Any some kind of robot. And kind of robot whom I choose to be made is mobile robot. Robot who is designed is avoider robot with addition aplication to dim the fire. The reason made this robot because in large scale this robot possible be involved in fire-frontier. Because this robot can look for the fire source. The architecture of avoider robot is shown in figure 1.

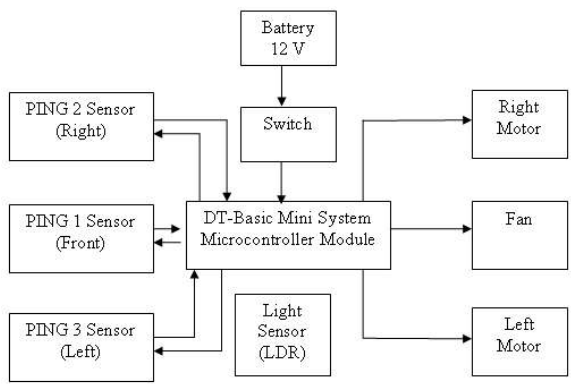

Figure 1. Avoider Robot Diagram Block

## 2. THEORY PRINCIPLE

Be explained some theory pronciple who have connected with avoider robot.

### 2.1 Distance Sensor

In this moment, distance sensor is sensor who is ussually used people to make robot. Many trademark of distance sensor which are sold in market. Like : distance sensor PING from Parallax, Devantech SRF04, GP2D02 from Sharp etc,. The reason this robot is made with distance sensor PING because this sensor easy to get in market and have price relatively more cheap than the other [1].

This sensor have ability to measure distance maximum about 3 m. With speed of burst on air about 340 m/s [2].

### 2.2 Light Sensor

Light sensor is mean sensor which work with measure light intensity which come. And light sensor which is used to make this robot is LDR (Light Dependent Resistor). With the reason cheap price. LDR is made from cadmium sulfida which sensitive from the light [3}.

### 2.3 Servo Motor

To movement, this robot use 2 servo motor on the right and left. Servo motor which be used is continuous servo motor form Parallax. This continuous servo motor have 360º rotation. Whereas which standard servo motor only have 180º rotation.

### 2.4 Comparator

Like which already explained, this robot use LDR sensor which is used to looking for the light (at experiment use candle). Comparator is used to compare the voltage of LDR with reference voltage which is given on the robot. For that, IC LM324 is needed.

### 2.5 Microcontroller

This robot is made with basic stamp 2 microcontroller which can be found at dt-basic mini system module. This microcontroller have ability more well than other microcontroller. The function of microcontroller is to control the robot. To make listing program robot be used software basic stamp editor which usually is given when buy dt-basic mini system module. Flowchart from listing program robot can be seen in figure 2.

## 3. ROBOTIC PROJECT

### 3.1 Basic Concept

Basicly, this robot is made so that the robot can be avoid block in around the robot. But, because other case this robot is added addition aplication like to dim the fire.

Like which explained before, listing program robot be written at basic stamp editor software. And this listing can be described like flowchart at below.

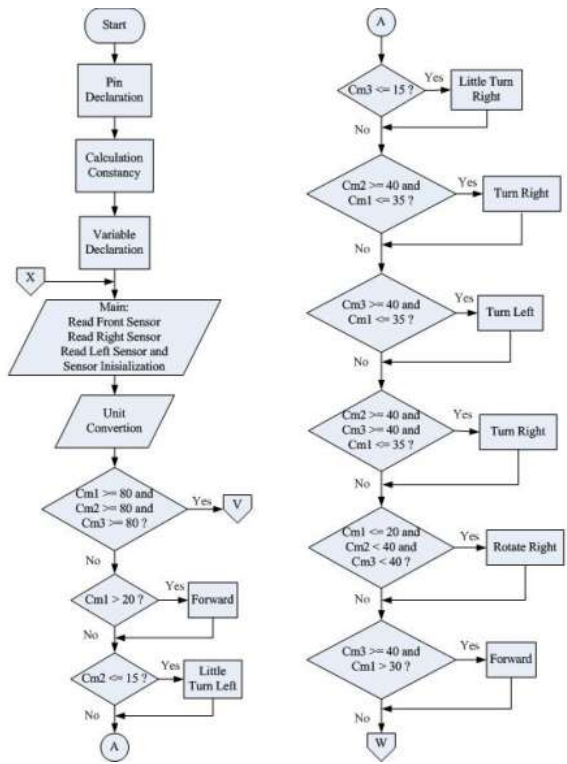

Figure 2a. Avoider Robot Flowchart

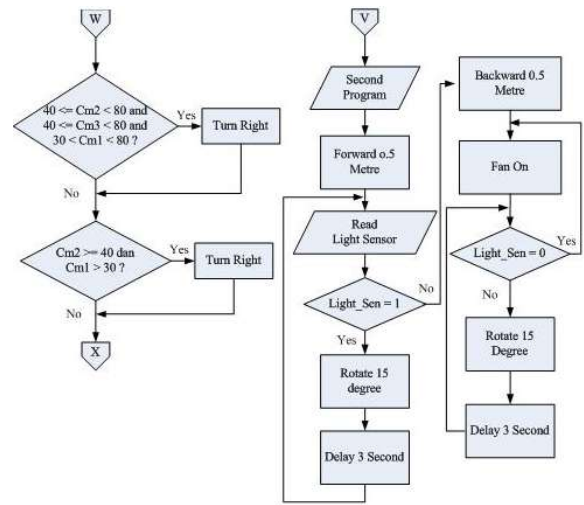

Figure 2b. Continuation Avoider Robot Flowhart

### 3.2 Planned of Lighting Sensor

Planned of lighting sensor can be seen in figure 3.

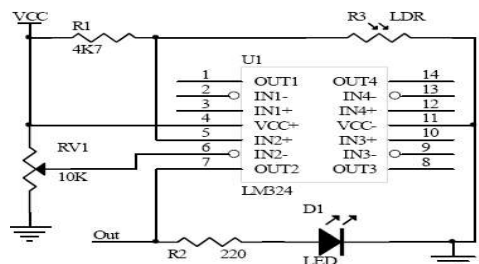

Figure 3. Planned of Lighting Sensor [4}

Planned of Lighting Sensor can be explined like this. If LDR sensor get light, resistance at LDR sensor will be down so that, LDR Voltage become small. Then, if LDR voltage is smaller than reference Voltage at robot so that the output will be low. Then, the output is low, robot will approach fire source. And then, when LDR voltage is bigger than reference voltage so that robot will try to look for the fire source with doing rotation the body until robot get its.

### 3.3 Movement

For movement, robot use servo motor. If robot will walk forward, right servo motor is given pulse so that can rotate CW (Clock Wise) and the left servo motor is given pulse so that can rotate CCW (Counter Clock Wise). And that case make robot can walk forward. This case possible to do if servo motor is given pulse 1000 ms for CW and 3500 ms for CCW.

### 3.4 Fan

This robot use fan which be usually used at computer casing. Function fan is to dim the fire.

## 4. RESULTS AND EXPERIMENT

we have presented result of avoider robot which we have made. The shape of avoider robot is shown at figure 4.

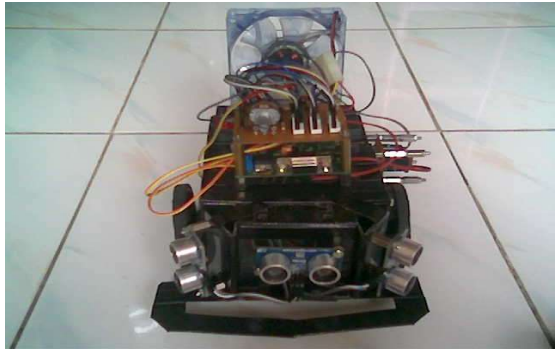

Figure 4. Shape of avoider robot.

To get best result, we done the experiment. Two experiment which done in these. First, done experiment to test ability each PING. And the second, experiment got output from planned of lighting sensor.

### 4.1 PING Experiment

In PING experiment be also founded two experiment. The first, to test ability aech pING, and the second, to get voltage value output PING with different condition.

#### 4.1.1 Test 1

In microcontroller, only content listing program to PING. And the result.

Table 1. Test 1 PING

| Distance (cm) | Ping 1 (V) (Front Sensor) | Ping 2 (V) (Right Sensor) | Ping 3 (V) (Left Sensor) |
|---|---|---|---|
| 10 | 0.14 | 0.15 | 0.16 |
| 20 | 0.16 | 0.18 | 0.19 |
| 30 | 0.18 | 0.19 | 0.20 |
| 40 | 0.20 | 0.21 | 0.22 |
| 50 | 0.23 | 0.24 | 0.24 |
| 80 | 0.30 | 0.30 | 0.30 |
| 100 | 0.34 | 0.36 | 0.36 |
| 200 | 0.62 | 0.61 | 0.58 |

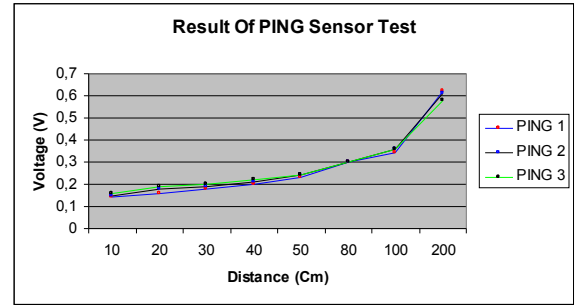

Figure 5. Curve of Test 1 PING

This case, can be explined like this. Because PING sensor have constant speed to burst travels and constant frequency so that if distance from PING and the object is more far, PING voltage is also increase. The result of test is shown in table 1.

#### 4.1.2 Test 2

In second test, microcontroller already content with listing program robot. In this second test, robot will meet different situation. Like turn right, turn left, t-juntion etc,. All of them be done to make easy when making maximum listing program from robot.

### 4.2 Planned of Lighting Sensor Test

For command the robot to dim fire especially robot must able to get fire source. If robot can get fire source, indicator LED in planned of lighting sensor will be off. This case, make robot to move to approach the target and then dim the fire.

Experiment is done to get voltage LDR value, and then compare with reference voltage at the robot. From experiment which be already done. The result, can be seen in table 2.

Table 2. Planned of Lighting Sensor Test

| Potensio Value (Ω) | Distance (Cm) | $V_{LDR}$ (V) | $V_{Ref}$ (V) | $V_{Out}$ (V) | Condition Of LED |
|---|---|---|---|---|---|
| 650 | 10 | 0.85 | 1.47 | 0.12 | Off |
| 650 | 20 | 1.43 | 1.46 | 0.13 | Off |
| 650 | 30 | 1.98 | 1.49 | 3.50 | On |
| 650 | 40 | 2.41 | 1.48 | 3.50 | On |
| 650 | 50 | 2.66 | 1.47 | 3.49 | On |
| 650 | 80 | 3.26 | 1.48 | 3.50 | On |
| 650 | 100 | 3.43 | 1.48 | 3.50 | On |

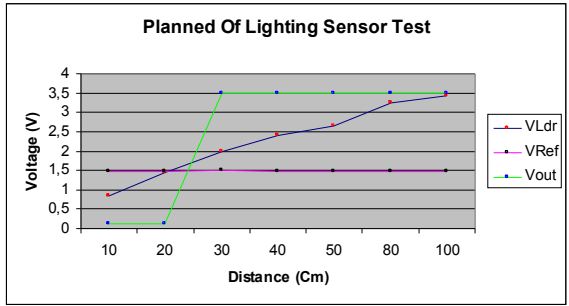

Figure 6. Curve of Planned of Lighting Sensor Test

## 5. CONCLUSION

- This robot able to avoid the block in around. This case, because PING sensor can be worked well.
- Although use LDR sensor, this robot already able to get fire source with measure light intensity which come.
- Color of the block is not influence for robot